\documentclass[3p,a4paper]{elsarticle}
\usepackage[english]{babel}
\usepackage[utf8]{inputenc}
\usepackage[cmex10]{amsmath}
\usepackage{lineno,hyperref}
\usepackage{xcolor}
\usepackage{graphicx}
\usepackage{array}
\usepackage{mdwmath}
\usepackage{mdwtab}
\usepackage{eqparbox}
\usepackage{pdflscape}
\usepackage{pbox}
\usepackage{color}
\usepackage{nccmath}
\usepackage{amssymb}
\usepackage{multirow}
\usepackage{subfig}
\usepackage{lscape}
\usepackage{setspace}
\usepackage{todonotes}

\modulolinenumbers[5]

\journal{Applied Soft Computing, Elsevier}

\newcommand{\fset}{A_j}
\newcommand{\lvar}{\Tilde{A}}
\newcommand{\model}{\mathcal{M}}

\begin{document}

\begin{frontmatter}

\title{Forecasting in Non-stationary Environments with Fuzzy Time Series}

\author[ifnmg,minds]{Petr\^onio C\^andido de Lima e Silva}
   \ead{petronio.candido@ifnmg.edu.br}
\author[ifmg,minds]{Carlos Alberto Severiano Junior}
    \ead{carlos.junior@ifmg.edu.br}
\author[minds]{Marcos Antonio Alves}
   \ead{marcosalves@ufmg.br}
\author[ufop,minds]{\\ Rodrigo Silva}
   \ead{rodrigo.silva@ufop.edu.br}
\author[braude,minds]{Miri Weiss Cohen}
   \ead{miri@braude.ac.il}
\author[dee,minds]{Frederico Gadelha Guimar\~aes\corref{mycorrespondingauthor}}
   \cortext[mycorrespondingauthor]{Corresponding author}
   \ead{fredericoguimaraes@ufmg.br}
   \ead[url]{https://minds.eng.ufmg.br/}
 
\address[minds]{Machine Intelligence and Data Science (MINDS) Laboratory, Federal University of Minas Gerais, Belo Horizonte, Brazil}
\address[ifnmg]{Federal Institute of Education Science and Technology of Northern Minas Gerais, Janu\'aria Campus, Brazil}
\address[ifmg]{Federal Institute of Education Science and Technology of Minas Gerais, Sabar\'a, Brazil}
\address[ufop]{Department of Computer Science, Federal University of Ouro Preto, Ouro Preto, Brazil}
\address[braude]{Department of Software Engineering, Braude College of Engineering, Karmiel, Israel}
\address[dee]{Department of Electrical Engineering, Universidade Federal de Minas Gerais, Belo Horizonte, Brazil}

\begin{abstract}
\begin{singlespace}
\noindent In this paper we introduce a Non-Stationary Fuzzy Time Series (NSFTS) method with time varying parameters adapted from the distribution of the data. In this approach, we employ Non-Stationary Fuzzy Sets, in which perturbation functions are used to adapt the membership function parameters in the knowledge base in response to statistical changes in the time series. The proposed method is capable of dynamically adapting its fuzzy sets to reflect the changes in the stochastic process based on the residual errors, without the need to retraining the model. This method can handle non-stationary and heteroskedastic data as well as scenarios with concept-drift. The proposed approach allows the model to be trained only once and remain useful long after while keeping reasonable accuracy. The flexibility of the method by means of computational experiments was tested with eight synthetic non-stationary time series data with several kinds of concept drifts, four real market indices (Dow Jones, NASDAQ, SP500 and TAIEX), three real FOREX pairs (EUR-USD, EUR-GBP, GBP-USD), and two real cryptocoins exchange rates (Bitcoin-USD and Ethereum-USD). As competitor models the Time Variant fuzzy time series and the Incremental Ensemble were used, these are two of the major approaches for handling non-stationary data sets. Non-parametric tests are employed to check the significance of the results. The proposed method shows resilience to concept drift, by adapting parameters of the model, while preserving the symbolic structure of the knowledge base. 
\end{singlespace}

\end{abstract}


\end{frontmatter}


\section{Introduction}
\label{sec:introduction}

The expanding utilization of smart sensors, the increasing availability of data storage, and the emergence of big data have led to an increasing amount of data being produced very often in the form of a stream \cite{cohem2008real,krawczyk2017ensemble,maia2020evolving}. In many real-world applications, this data is organized in the form of a time series. In a time series forecasting problem, the information available for the prediction is limited to the past values of the series \cite{cohem2008real}. Hence, the temporal relationships which describe the evolution of the series must be deduced exclusively from these values.

Generally, the characteristics of the processes which generate a time series are unknown \cite{assaad2008timeseries}. In several cases of practical interest, such as stock indices \cite{qiu2019fusion} in finance, evapotranspiration in agriculture \cite{zhao2019ensemble}, among others, the variable of interest in the time series is in fact the fusion of many other data sources. Not by accident these kinds of time series tend to show highly non-linear and non-stationary patterns \cite{qiu2019fusion,zhao2019ensemble}.

Many forecasting methods assume that the data is generated from a fixed probability distribution. However, as mentioned before, many time-series related applications deal with non-stationary and heteroskedastic stochastic processes which may arise from phenomena such as: seasonality, periodicity, hardware and machine faults, aging sensors and components, and unexpected events. These changes modify the properties of the data generating process, then changing its underlying probability distribution over time. In such non-stationary environments, any non-adaptive model trained under the false stationarity assumption is deemed to present a progressively increasing error or simply fail at some point \citep{Ditzler2015}. 

Fuzzy Time Series (FTS) was introduced by Song and Chissom in 1993 \citep{song1993fuzzy} to handle with vague and imprecise knowledge in time series data. In FTS, the domain of the variable of interest, called Universe of Discourse (UoD), is divided into sub-domains, and each of them is linked to a fuzzy set. After the construction of these fuzzy sets, temporal patterns of the type IF-THEN are extracted from the training data in order to identify a rule-base able to represent the generating function of the time series.

FTS forecasting methods have become attractive due to their simplicity, model transparency, forecasting accuracy and computational performance. Some examples of successful applications are found in tourism demand forecasting \citep{lee2011weighted}, energy load \citep{chen2016electric,sadaei2017short,severiano2017very}, stock index price predictions \citep{Chen2011,silva2016interval, talarposhti2016stock}, and many more. However, when dealing with non-stationary stochastic processes, the values of the time series might go outside the UoD as defined from the training data. Furthermore, the initial setting of the fuzzy sets may become inadequate over time due to the lack of mechanisms that will allow the membership functions to adapt to the varying behavior of the time series \citep{garibaldi2008nonstationary}. 

In \citep{song1993fuzzy}, Song and Chisom presented an approach to induce time-variant Fuzzy Time Series, by retraining the FTS model in a sliding window. Thus, every time a new data point is fed to the model, the FTS has to be retrained from scratch. In many cases, this requirement may render the methods impractical due to the related computational costs. 

Recently, in \citep{2018esann}, a first attempt was made to avoid retraining in FTS when they are applied to non-stationary environments. In this approach, Non-Stationary Fuzzy Sets (NSFS) \citep{garibaldi2008nonstationary} were employed to forecast heteroskedastic time series with unconditional variance, i.e., time series where the variance changes through time in a predictable way. Their approach, however, is not able to account for conditional variances and scenarios with concept-drift.


Given the need to produce adaptive models for forecasting in non-stationary environments \cite{bose2019designing}, in this paper, we introduce a Non-Stationary Fuzzy Times Series (NSFTS) method with time varying parameters adapted from the data distribution. 

In the proposed approach, the FTS is also built on Non-Stationary Fuzzy Sets \citep{garibaldi2008nonstationary}. Based on the residual errors, different perturbation mechanisms adapt the membership functions in response to statistical changes in the time series. Thus, the fuzzy sets will reflect changes in the stochastic data generating process without model retraining. Differently from \citep{2018esann}, the proposed mechanisms give to the FTS the ability to handle non-stationary and heteroskedastic data as well as scenarios with concept-drift.

To validate the proposal, different data sets consisting of market indices, FOREX pairs, cryptocurrency exchange rates and synthetic data were used. These data sets were selected because they are all non-stationary and present different types of concept drift. The forecast accuracy of the proposed method was also compared with other methods.

The remainder of this work is organized as follows: in Section \ref{sec:preliminaries}, the main concepts of Fuzzy Time Series and non-stationary Fuzzy Sets are introduced; in Section \ref{sec:timevariant}, time variant methods are discussed; in Section \ref{sec:methodology}, the NSFTS method is presented and Section \ref{sec:results} discusses the results of the computational experiments performed to compare the proposed method against the others methods. Finally, in Section \ref{sec:conclusion}, the main findings of the research are synthesized. 


\section{Preliminaries}
\label{sec:preliminaries}

\subsection{Non-stationary Time Series}

Time series data observed in different real-world applications are often non-stationary.
Given that a stationary time series is defined in terms of its mean and variance, non-stationarity can be detected if any (or both) of these components vary over time. Thus, in a non-stationary context, if the chosen forecasting model relies on a false assumption of stationarity, there is a significant risk that it will present a degrading performance over time and eventually become obsolete \citep{Ditzler2015}.

Some non-stationary time series are consequence of hidden contexts, not given in the form of predictive features \citep{Tsymbal2004}. In such situations, inferring a forecasting model becomes more difficult, since changes in the hidden context can induce unpredictable changes in the target concept. These changes are known as \textit{concept drift}  and since they usually cannot be identified explicitly \cite{Ditzler2015}, an effective forecasting model should be able to quickly adapt to the resulting changes in the time series data.

A symptom of concept drift in a time series is the change of the parameters that define its distribution \citep{gama2014survey}. Following \citep{gama2014survey}, the change in the distribution can be classified with respect to the rate at which the drift occurs. For instance, a political event that suddenly causes a strong effect on the stock market is a case of \textit{abrupt concept drift} observed in a time series. Another example is the aging effects of a sensor which gradually leads to lower performance in a device. Such case can be referred to as \textit{gradual concept drift}. According to Ditzler et al. \citep{Ditzler2015}, the drifts, whether abrupt or gradual, can also be classified as permanent or transient. The former is related to the effect of variation, and is not limited in time, while the latter occurs in a limited time window followed by an effect of disappearance.

Due to the practical importance of problems with varying statistical properties, the literature has presented some alternatives to handle the issue. Popular approaches commonly used in forecasting methods, including FTS, are: (i) detrending  \citep{kim1999fuzzy,huang2003applications}, in which a data transformation is applied to the time-series to remove the trend component and (ii) time-varying models derived from the distribution of the raw data \citep{liu2017simulation}. 

\subsection{Fuzzy Time Series}

\begin{table}[!ht]
    \centering
    \begin{tabular}{c|l} 
       Symbol & Description  \\ \hline 
         $\Omega$ & Fuzzy time series order (lags) \\ 
         $k$ & Number of fuzzy sets for partitioning the universe of discourse\\
         $\mathcal{M}$ & Time series model \\
         $t$ &  Time instant\\
         $y(t)$ & Time series value at time $t$\\
         $\hat{y}(t)$ & Estimated time series value at time $t$\\
         $U$ & Universe of discourse \\
         $A_j$ & j-th fuzzy set\\
         $\mu_{A_j}$ & Membership function of fuzzy set $A_j$\\
         $T$ & Number of time steps\\
         $f(t)$ & Fuzzified value of the series at time $t$\\
         $c$ & Fuzzy set midpoint \\
         $W$ & Window of observations\\
         $\pi(\cdot)$ & Perturbation function for the non-stationary fuzzy sets\\
         $\mathcal{E}$ & Set of residuals\\
         $\lvar$ & Linguistic variable \\
         $R$ & Refreshing interval \\
         $\delta$   & Displacement applied to the fuzzy set  \\
         $\rho$     & Scaling factor  \\
         $w$    & Window size  \\
         $n$    & Sample size  \\ %
         $LHS$ & Set of fuzzy sets in the left hand side of a fuzzy logical relationship\\
         $RHS$ & Set of fuzzy sets in the right hand side of a fuzzy logical relationship\\
    \end{tabular}
    \caption{Convention of symbols}
    \label{tab:symbols}
\end{table}

Since the introduction of the Fuzzy Time Series in \citep{song1993fuzzy}, several categories of FTS methods have been proposed, varying mainly by their order $\Omega$, number of fuzzy sets $k$ and time-variance \citep{bose2019designing} -- see Table \ref{tab:symbols} for the convention of symbols adopted here. The order is defined by the number $\Omega$ of time-delays (lags) that are used in modeling the time series. The time variance defines whether the FTS model changes over time, with the Time Invariant models $\mathcal{M}^t$ having the same parameters for all $t=0, \hdots, T$, and the Time Variant models $\mathcal{M}^t$ having different parameters in different time instants $t$. 

Given an univariate time series $Y \in  \mathbb{R}$ , where $y(t) \in Y$ are the instances of $Y$ for $t = 0, 1, \hdots, T$, the UoD is limited by the known bounds of $Y$, such that $U = [\min(Y),\max(Y)]$. The training procedure of an FTS model $\mathcal{M}$ consists of the following three steps: 
 \begin{itemize}
     \item[a)] Partitioning: split $U$ in $k$ overlapping intervals. For each interval a new fuzzy set $\fset$ is created, each one with its own membership function (MF) $\mu_{\fset}$. A linguistic value $\lvar$ is assigned to each fuzzy set and represents a region of $U$. The computational cost of this step is $O(k)$.
     
     \item[b)] Fuzzification: maps the crisp time series $Y$ onto the fuzzified time series $F$, by replacing each $y(t) \in Y$ by the fuzzified value $f(t) = \mu_{\fset}(y(t))$, $\forall \fset \in \lvar$, for $t = 1, \hdots,T$. The computational cost of this step is $O(T \cdot k)$.
     
     \item[c)] Knowledge Extraction: creates a representation of the sequential patterns in the time series. In a rule-based FTS, as in \citep{chen1996forecasting}, the rules have a format $A_i^{t-\Omega},\ldots,A_i^{t-1} \rightarrow A_j, A_k,\ldots$, where  $A_i^{t-\Omega},\ldots,A_i^{t-1} \in \lvar$ is the precedent and $A_j,A_k,\ldots \in \lvar$ is the consequent. The rule can be read as ``IF $y(t-\Omega)$ is $A_i^{t-\Omega}$ AND $\ldots$ AND $y(t-1)$ is $A_i^{t-1}$ THEN $y(t)$ may be $A_j, A_k,\ldots$''. The computational cost of this step is $O(T\cdot k^\Omega)$.
 \end{itemize}

Once the model $\model$ is trained it can be used to forecast values of $Y$ given a sample $y(t-\Omega),\ldots,y(t-1)$ with a three step procedure:

\begin{itemize}
     \item[a)] Fuzzification: maps the crisp sample $y(t-\Omega),\ldots,y(t-1)$ onto the fuzzified values $f(t-\Omega),\ldots,f(t-1)$, where each $f(t) = \mu_{\fset}(y(t))$, $\forall \fset \in \lvar$, for $t = \Omega.\ldots,1$. The computational cost of this step is $O(\Omega\cdot k)$.
     
     \item[b)] Rule Matching: find in $\model$ the rules whose the precedent matches with the fuzzified values $f(t-\Omega),\ldots,f(t-1)$. The activation $\mu_j$ of each rule $j$ is the minimum T-norm of the individual membership values of each fuzzified value. The computational cost of this step depends on the length of the sample ($\Omega$) and the number of rules in $\model$ ($k^\Omega$), then the complexity is $O(\Omega \cdot k^\Omega)$.
     
     \item[c)] Defuzzification: the estimated value of $\hat{y}(t)$ is calculated by finding the mean value $c_j$ of each matched rule $j$, by averaging the midpoints of the consequent fuzzy sets, and then calculating the sum of the mean values of each rule weighted by its activation values 
     \[
     \hat{y}(t) = \frac{\sum_{j} \mu_j\cdot c_j}{\sum_{j} \mu_j}
     \]
 \end{itemize}

Many improvements were proposed in the FTS literature. The High-Order FTS (HOFTS) \citep{severiano2017very} extended the classical method in \citep{chen1996forecasting} by using several lags in the forecast and it is able to recognize more complex patterns in the time series. In 
\cite{yu2005weighted}, 
\cite{Cheng2008} and 
\cite{sadaei2014short} weights are added in the consequent of the rules in order to give more importance for certain fuzzy sets. More recently, the Probabilistic Weighted FTS (PWFTS) was proposed in \citep{Silva2019b}, and made available in the pyFTS library \cite{pyFTS}, including weights in both precedent and consequent of the FTS rules, achieving high accuracy and outperforming traditional forecasting approaches.

In order to represent forecasting uncertainty, \citep{silva2016interval}, \citep{Silva2019b} and \citep{Silva2017ensemble} proposed approaches for probabilistic forecasting. In \citep{Silva2019} and \citep{Silva2019a} distributed FTS variants are proposed for Big Data time series. Multivariate time series are explored in \citep{Silva2019c}, which uses Fuzzy Information Granules (FIG) to propose a multivariate forecasting method. In \citep{bose2019designing} a survey on the design of FTS forecasting models was provided. A comprehensive review of those aforementioned models, focused on time-invariant and rule-based approaches can be found in \citep{Silva2019a}. 

The major hyper-parameters of FTS methods are the number of fuzzy sets $k$ and the order of the model $\Omega$. These hyper-parameters conduce the model training and forecasting and are responsible for the accuracy and model parsimony (the number of rules). A method for hyperparameter tuning of FTS models is presented in \cite{patricia2020}.

The FTS approaches listed in this section are all time-invariant approaches which means that the  models are trained only once and then its internal rule base does not change. To keep their accuracy, these models must make strong assumptions about the stationarity and homoskedasticity of the time series. In non-stationary environments, this is a major drawback.

\begin{figure*}[!htb]
\centering
\includegraphics[width=1\textwidth]{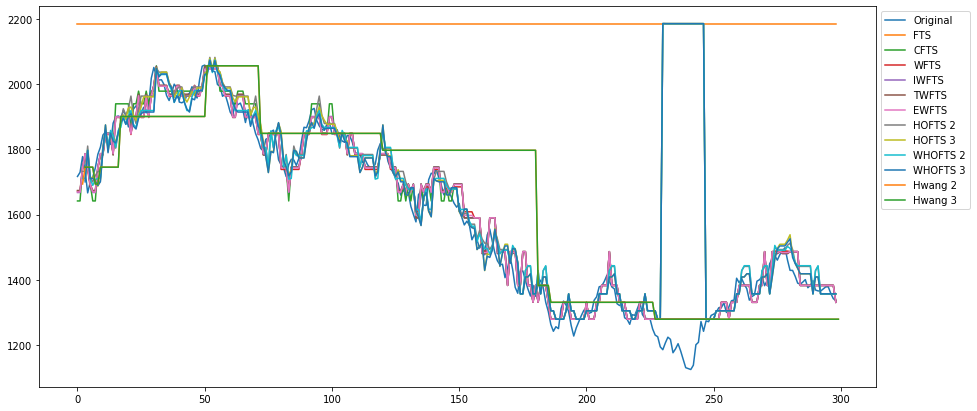}
\caption{Sample of models forecasts in a concept drift scenario. The dataset (Original) is NASDAQ. The models are: Fuzzy Time Series (FTS) \cite{song1993fuzzy};
Conventional FTS (CFTS) \cite{chen1996forecasting};
Weighted FTS (FTS) \cite{yu2005weighted};
Improved Weight FTS (IWFTS) \cite{efendi2013iwfts};
Trend-Weighted FTS (TWFTS) \cite{cheng2006twfts};
Exponentially Weighted FTS (EWFTS) \cite{sadaei2014short};
High Order FTS (HOFTS) \cite{severiano2017very};
Weighted High Order FTS (WHOFTS) order 2 and 3 \cite{Silva2019a};
Hwang \cite{hwang1998handling} order 2 and 3.}
\label{fig:models_plot}
\end{figure*}

To better understand the behavior of the FTS methods in non-stationary scenarios, Figure \ref{fig:models_plot} shows the performances of several methods in the literature when the test data falls out of the known UoD. For this example, we used the NASDAQ dataset. It can be seen that this situation makes most of the trained models useless in the long run. 

\subsection{Non-stationary Fuzzy Sets}

Non-stationary fuzzy reasoning and non-stationary fuzzy sets (NSFS) were introduced by Garibaldi and Ozen and Garibaldi, Jaroszewski and Musikasuwan, respectivelly in \citep{garibaldi2007uncertainmaking} and \citep{garibaldi2008nonstationary}. The main idea is to extend the traditional fuzzy set definition by introducing a dynamic component that changes the membership function $\mu$ over time.  This change takes several forms: a) variation in location: by displacing the parameters of the $\mu$ function along the UoD without changing its shape; b) variation in width: changing the shape of $\mu$ by stretching or contracting its bounds; and c) noise variation: by adding random noise to the membership grade. 

A NSFS is defined with two functions: the non-stationary membership function (NSMF), which considers time variations of the corresponding membership function (MF), and the perturbation function, which is the dynamic component responsible for altering the parameters of the membership function given some parameter set.

According to Garibaldi et al. \citep{garibaldi2008nonstationary} a non-stationary fuzzy set $\dot{A}$ can be formalized as follows:

\begin{equation} \label{eq_nsmf}
  \dot{A} = \int_{t \in T} \int_{x \in X} \mu_{\dot{A}}(t,x)dxdt 
\end{equation} 
where $\dot{A}$ is a fuzzy set over a UoD $X$ characterized by a NSMF $\mu_{\dot{A}}(t,x)$ at a set of time points $T$.

It is worth noting that the NSMF $\mu_{\dot{A}}$ varies throughout $U$ and the time interval $T$. This means that the NSMF parameter set should vary over time. A regular MF, $\mu_A(x)$, can be expressed as $\mu_A(x, p_1, \hdots, p_m)$, where $p_1, \hdots, p_m$ denote the parameters of $\mu_A(x)$. Thus, a NSMF can be denoted in an analogous way as follows:

\begin{equation} \label{eq_muatx}
    \mu_{\dot{A}}(t,x) = \mu_A(x,p_1(t), \hdots, p_m(t))
\end{equation}
where each parameter can be varied over time by a perturbation function multiplied by a constant.

One of the constraints existing in several FTS models is the lack of ability to deal with conditional variances and concept-drift scenarios, in which the statistical characteristics of the time series change, sometimes drastically. Thus, through small variation in the MFs, we deploy NSFS in FTS models 
to deal with variability for decisions over time and contributing to the mitigation of this drawback. 

\section{Time Variant Approaches}
\label{sec:timevariant}


Time-variant FTS models should be applied when the data is not compliant with the stationarity and homoskedasticity assumptions. They include incremental, flexible and evolving techniques for adapting the model to the input data \citep{bose2019designing,song1994forecasting,liu2010improved}. 

In the seminal work of Song and Chissom on time variant FTS \cite{song1994forecasting}, time-variance can be seen as a meta-modeling technique. It is not a proper FTS model, it is a training policy for another FTS method which controls when this method will be retrained and how many lags will be used. More specifically, the time variant approach defines $W$, the length of the memory window, and $R$, the refreshing interval. Thus, the chosen FTS model is built from scratch every $R$ time instants using the most recent $W$ observations of the time series. 


This first time-variant FTS approach  \citep{song1994forecasting} was followed by several other authors who have mixed its training policy with different knowledge models and weighting schemes \citep{liu2010improved,singh2007simple,jilani2008refined,vovan2018improved}. 

\begin{figure}[htb]
    \centering
    \includegraphics[width=1\textwidth]{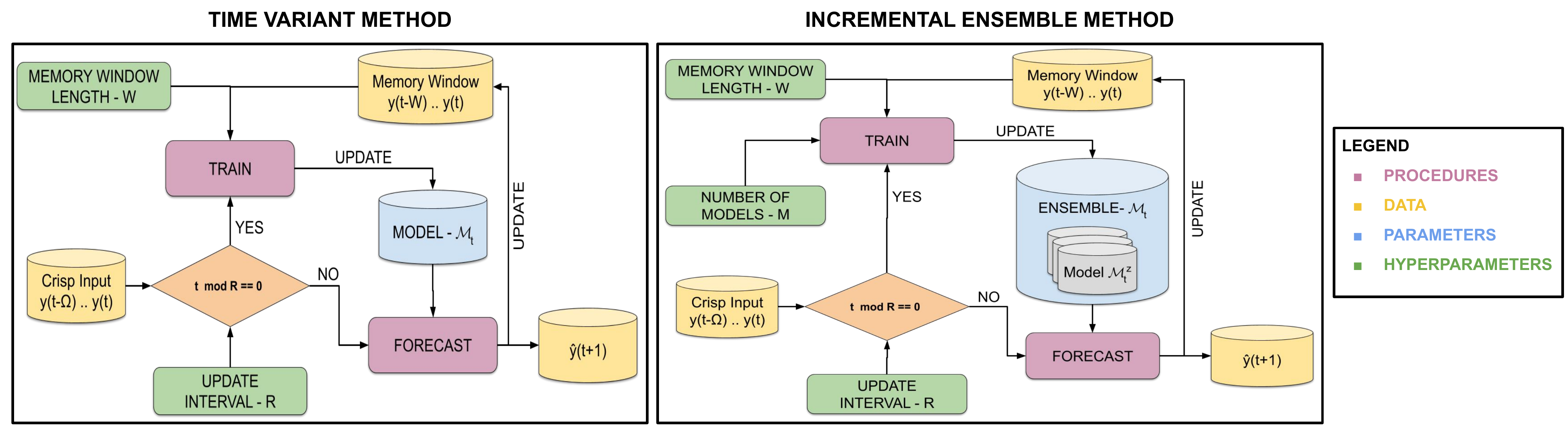}
    \caption{Song and Chissom method and Incremental Ensemble}
    \label{fig:time_variant}
\end{figure}

In Figure~\ref{fig:time_variant} (left) it is easy to see that all FTS time invariant approaches can be combined with the time variant method and used as the internal model. However, two major drawbacks of the classical Song and Chissom's method can be highlighted. The first one is its limited memory. Once a new data point arrives the previous knowledge base is completely discarded. This, in turn, may lead to catastrophic forgetting of frequent patterns if the parameters $W$ and $R$ are not tuned correctly. The second one is its high computational cost. Using a binary search tree structure to organize the $k$ fuzzy sets, the time complexity for a search among them decreases from $O(k)$ to $O(\log k)$. Thus, for a given input of size $T$, the complexity is $O(T/R\cdot W \cdot (\log k)^\Omega)$, given that its internal model will be retrained $T/R$ times with a potential cost of $O(W \cdot (\log k)^\Omega)$.

The Incremental Ensemble, see Figure \ref{fig:time_variant} (right), is an alternative to control the limited memory of the Song and Chissom method by balancing the learning of new behaviors with the memory of the old ones \cite{krawczyk2017ensemble,junior2019iterative,assaad2008timeseries,gama2014survey}. The Incremental Ensemble is, in fact, a meta-model containing $M$ internal models. It is also controlled by the $W$ and $R$ parameters. At each interval of $R$ observations a new model $\model_t$ is built with the last $W$ observations. $\model_t$ is then appended to the ensemble while the oldest model $\model_{t-M}$ is discarded. 

The generalization of Song and Chissom's method as well as the Incremental Ensemble using FTS as internal models can be seen in Figure \ref{fig:time_variant}. As in the Song and Chissom's case, the major drawback of the incremental ensemble techniques is its computational cost. 

\section{The Non-Stationary Fuzzy Time Series method}
\label{sec:methodology}


The proposed Non-Stationary Fuzzy Time Series method extends the concepts of the Conventional FTS method \citep{chen1996forecasting} to incorporate Non-Stationary Fuzzy Sets presented by Garibaldi et al. \citep{garibaldi2008nonstationary}. In the proposed forecasting procedure, the mean and variance of the residuals are used to translate and scale the sets to adapt them to the changes occurred in the data after the training process. The parameters of the fuzzy sets change towards canceling the mean of the residuals. If the variance of residuals is high, the range of the fuzzy sets should be reduced to reduce granularity. Moreover, if the bounds of U change, the sets are adapted to respond to this change as well.

By convention, it is assumed that $Y \in \mathbb{R}$ is a real valuated univariate time series and $y(t) \in Y$ a single data point indexed by time index $t \in \mathbb{N}$. It is also assumed that all fuzzy sets have a membership function, $\mu$, with triangular shapes, as defined in equation \eqref{eqn:membership}, where $x\in Y$ is the input value and $l,c,u \in Y$ are, respectively, the lower, midpoint and the upper basis of the triangle.

\begin{equation}
\mu(x,l,c,u) = \left\{ \begin{array}{lcr}
     0 & if & x < l \; or \; x > u \\
     \frac{x - l}{c - l} & if & l \leq x \leq c \\
     \frac{u - x}{u - c} & if & c \leq x \leq u 
\end{array}\right.
\label{eqn:membership}
\end{equation}

\begin{figure*}[!htb]
    \centering
    \includegraphics[width=\textwidth]{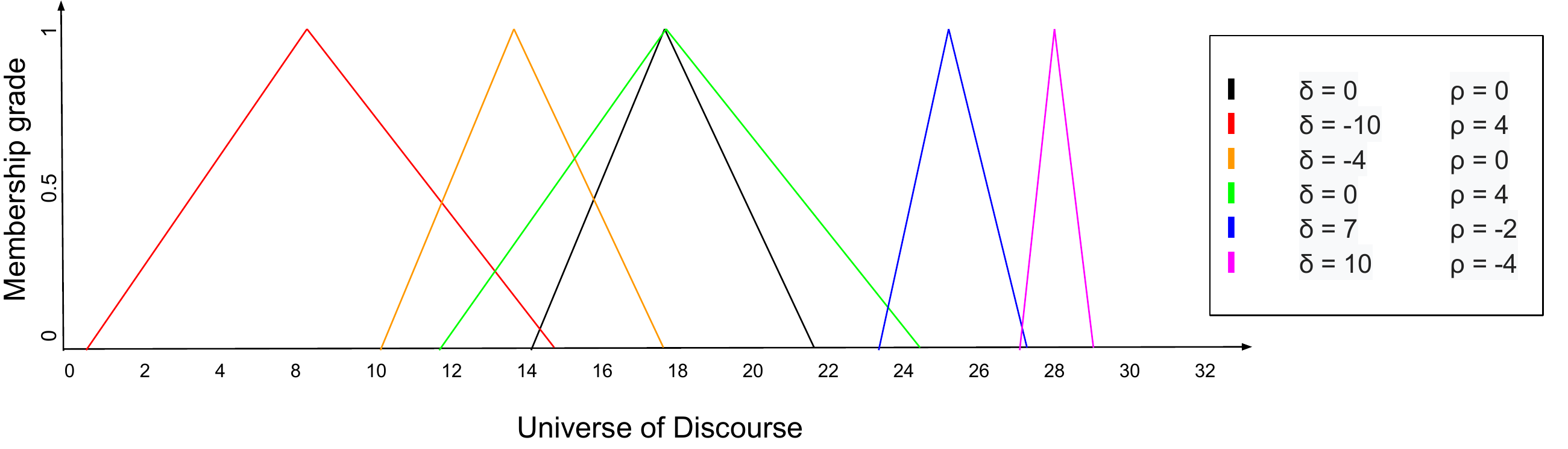}
    \caption{The effects of the $\delta$ and $\rho$ perturbation parameters on a triangular membership function with $l=14$, $c=18$, $u=22$}
    \label{fig:nsfs}
\end{figure*}

Furthermore, all NSFS 
have a perturbation function $\pi$, defined in  \eqref{eqn:nsfs}, over the parameter set $\{l, c, u\}$ of $\mu$,  where $\delta \in \mathbb{R}$ is the displacement and $\rho \in \mathbb{R}$ is the scale increment. $\pi$ shifts the triangular function across the domain of $Y$ and scales it by moving the triangle bounds with respect to the center as shown in Figure \ref{fig:nsfs}. Thus, the NSMF 
function is given by:

\begin{equation}
\mu(x, \pi(l,c,u,\delta,\rho))    
\end{equation}
\noindent where,
\begin{equation} 
\label{eqn:nsfs}
\pi(l,c,u,\delta,\rho) = \left\{ \frac{\rho}{2}-(l+\delta)\;,\; c+\delta\;,\; \frac{\rho}{2}+(u+\delta) \right\}
\end{equation}

\begin{figure}[!htb]
    \centering
    \includegraphics[width=0.6\textwidth]{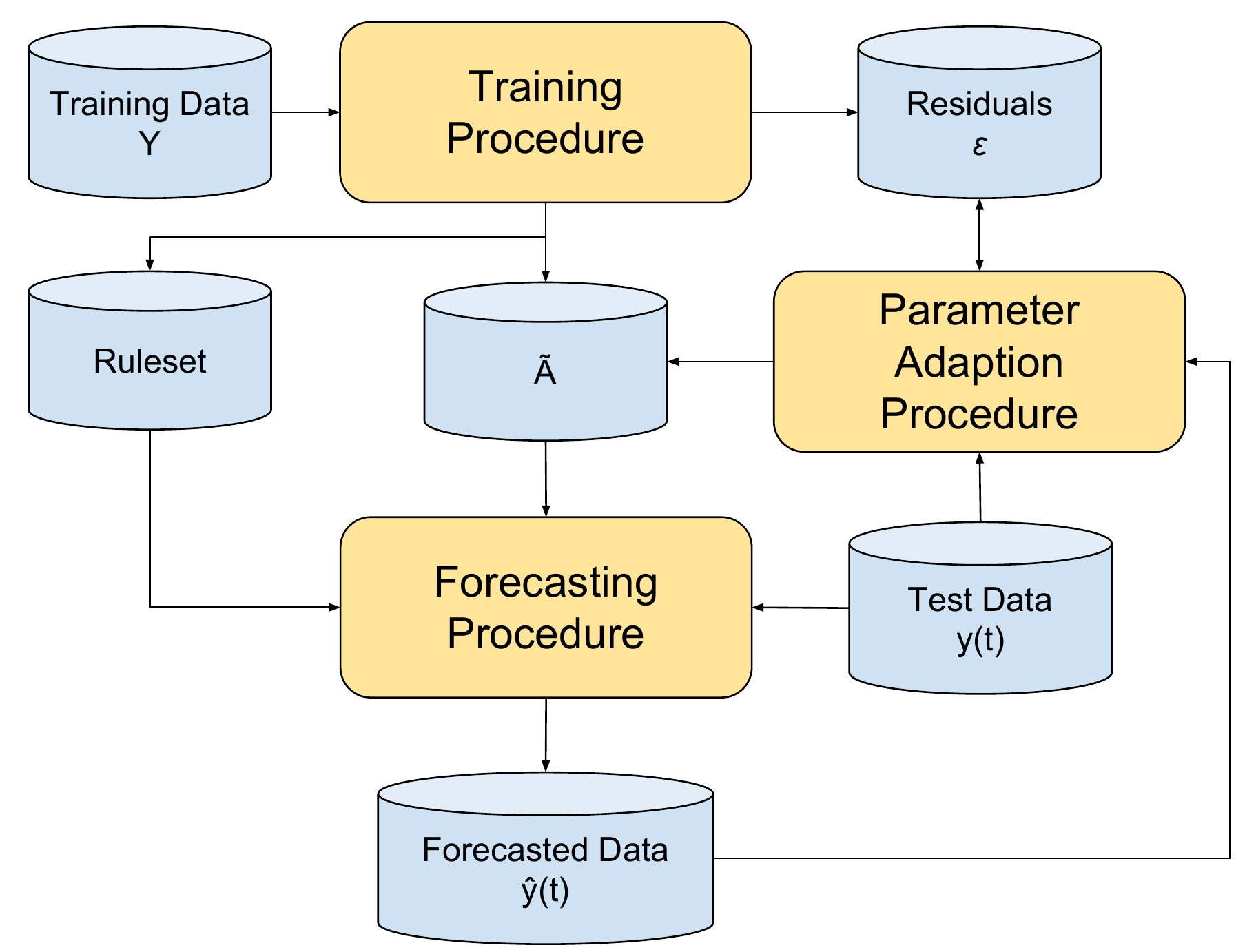}
    \caption{NSFTS overview}
    \label{fig:nsfts_all}
\end{figure}

The proposed method, depicted in Figure \ref{fig:nsfts_all}, consists of training, parameter adaptation and forecasting procedures. 

In the training procedure, the $U$ partitioning creates static fuzzy sets. In a forecasting model, a normal distribution of the residuals suggests that its predictive ability is consistent over the entire data range. Therefore the objective of the training procedure is to generate a model that captures all the information in the data, leaving a residual $\mathcal{E} \sim N(0,1)$. This is done by defining  the $l$, $c$ and $u$ parameters for each one of the $k$ fuzzy sets and using them to extract temporal patterns from the data. The $w$ most recent residuals of the model are stored in the set $\mathcal{E}$ which consists of the differences between the model forecasts and the new data collected in that window.


Given a non-stationary scenario, it is unlikely that a model with predefined and fixed parameters could manage to keep its residuals normally distributed. In the parameter adaption procedure, the mean and variance of the residuals are monitored and used to adapt the MFs. 
The NSFS is perturbed in order to keep the residuals as close to $\mathcal{E} \sim  N(0,1)$ as possible. The goal of this procedure, is to find the best values for the parameters $\delta$ and $\rho$ of each NSFS 
$\pi$ function in order to adapt them to the changes in the data.


The forecasting procedure finds the rules that match a given numerical input and use them to compute a numerical forecast using the non-stationary fuzzy sets perturbed by the parameters $\delta$ and $\rho$. 

It can be noticed that the rationale behind the proposed method is to keep the residuals normally distributed. In this context, the main technical contribution of this paper is to introduce the adaptation procedure which will keep $\mathcal{E} \sim  N(0,1)$. This procedure is detailed in section \ref{sec:parameter_adaption_procedure}. For completeness of presentation, the training and forescasting procedures defined in \citep{chen1996forecasting} are reproduced in subsections \ref{sec:training_procedure} and \ref{sec:forecasting_procedure}, respectively.


\subsection{Training Procedure}
\label{sec:training_procedure}

Given the training data, $Y$, the number of partitions, $k$, and the length of the residuals window, $w$:

\begin{enumerate}
\item [Step 1] \textit{Defining the universe of discourse, $U$}: 

\begin{equation}\label{eqn:uod}
U = [lb, ub]
\end{equation}

\noindent where, $lb = \min(Y)-\min(Y)\times 0.2$ and $ub = \max(Y) + \max(Y)\times 0.2$.

\noindent Notice that the data bounds are extrapolated to compensate for a possible underestimation of the bounds in the training set. The value $0.2$ is a typical value, but can be modified according to the problem. 

\item[Step 2] \textit{$U$ Partitioning}: Define the midpoints $c_i, i = 0,...,k-1$ of the initial fuzzy sets;

\begin{equation}
    c_i = lb + i \times \frac{(ub - lb)}{(k-1)} 
\end{equation}

\item[Step 3] \textit{Define the linguistic variable $\Tilde{A}$}: Create $k$ overlapping fuzzy sets $A_i$, with triangular MF $\mu_{A_i}$ as defined in equation \eqref{eqn:membership2}. 

\begin{equation}
\mu_{A_i}(x) = \left\{ \begin{array}{lcr}
     0 & if & x < l \; or \; x > u \\
     \frac{x - l_i}{c_i - l_i} & if & l_i \leq x \leq c_i \\
     \frac{u_i - x}{u_i - c_i} & if & c_i \leq x \leq u_i 
\end{array}\right.
\label{eqn:membership2}
\end{equation}

where $l_i = c_{i-1}$ and $u_i = c_{i+1}$.

Each fuzzy set $A_i \in \Tilde{A}$ is a linguistic term of the linguistic variable $\Tilde{A}$. Once the fuzzy sets are created, a function $\pi_i$, as defined in equation \eqref{eqn:nsfs}, will be associated with each fuzzy set in order to transform it into a NSFS, 
initialized with $\delta = 0$ and $\rho = 0$.

The number of sets $k$ defines the number of states and consequently the number of state transitions that the model can represent. The more complex the time series the greater the number of $k$ should be. One should, however, be careful to not overestimate $k$ since it may cause overfitting and make the model unnecessarily big.

\item[Step 4] \textit{Fuzzification}: Transform the original time series data $Y = \{y(0), y(1), \hdots, y(T)\}$ into a fuzzy time series $F = \{f(0), f(1), \hdots, f(T)\}$, where $f(t)$ is defined as: 

\begin{equation}
    f(t) = \{ \mu_{A_0}(y(t)),\mu_{A_1}(y(t)),...,\mu_{A_{k-1}}(y(t))\}
\end{equation}

\item[Step 5] \textit{Generate the temporal patterns}: The fuzzy temporal patterns have format $A_l \rightarrow A_r$, where:

\noindent The precedent (or Left Hand Side - LHS) is:
\begin{equation}
A_l = \underset{A_i}{\arg \max}\left ( \mu_{A_i}(y(t-1)) \right )
\end{equation}

\noindent And the consequent (or Right Hand Side - RHS) is:
\begin{equation}
A_r = \underset{A_i}{\arg \max}\left ( \mu_{A_i}(y(t)) \right )
\end{equation}

\item[Step 6] \textit{Generate the rule base}: Select all temporal patterns with the same precedent and group their consequent sets creating a rule with the format $A_l \rightarrow A_a, A_b,...$. Thus the rule $RHS$ can be understood as the set of possibilities which may happen on time $t+1$ (the consequent) when a certain set $A_l$ is identified on time $t$ (the precedent).

\item[Step 7] \textit{Compute the residuals $\epsilon$}: Invoke the Forecasting Procedure defined on Section \ref{sec:forecasting_procedure} using the given training data, $Y = \{ y(0),...,y(T)\}$, as the input and forecast the last $w$ items to calculate the set of residuals $\mathcal{E}$ defined as: 

\begin{equation}
    \mathcal{E} = \{ \epsilon(t-w), \epsilon(t-(w-1)),...,\epsilon(t) \}
\end{equation}

\noindent where $\epsilon(t) = y(t) - \hat{y}(t)$ and $\hat{y}(t)$ is the forecast produced by the model.

\end{enumerate}

\subsection{Parameter Adaption Procedure}
\label{sec:parameter_adaption_procedure}

Given an input value $y(t)$, the last forecast value $\hat{y}(t)$, the length of the residuals vector $w$, the residuals set  $\mathcal{E}$ and the linguistic variable $\Tilde{A}$, do:

\begin{enumerate}
\item [Step 1] \textit{Find the displacements of $y(t)$ on $U$}: If $y(t)$ is below the lower bound of $U$ store the difference in $d_l = lb-y(t)$, otherwise $d_l = 0$. If $y(t)$ is above the upper bound of $U$, store the difference in 
$d_u=y(t)-ub$, otherwise $d_u = 0$. Then compute the displacement range $r$ as the sum of the UoD displacements $d_u$ and $d_l$, according with \eqref{eqn:new_range}, and the displacement midpoint $mp_r$ as $r$ divided by 2, according with \eqref{eqn:range_midpoint}

\begin{equation}\label{eqn:new_range}
    r = d_u - d_l
\end{equation}

\begin{equation}\label{eqn:range_midpoint}
    mp_r = r/2
\end{equation}

\item [Step 2] \textit{Compute the mean $\overline{\mathcal{E}}$ and variance  $\sigma_\mathcal{E}$ of the set $\mathcal{E}$}: The residual mean $\overline{\mathcal{E}}$ indicates a bias, a shift on the accuracy of the trained model that must be corrected. The variance, $\sigma_\mathcal{E}$, represents the shift on the trained model variance, as a reflection to a change in the test data. These values will be used to adjust the position and length of the fuzzy sets.

\item [Step 3] \textit{Compute the displacements, $\delta_i$}: The  displacement is computed for each fuzzy set $A_i$ in order to equally move the $k$ fuzzy sets by the mean shift $\overline{\mathcal{E}}$, one fraction of the displacement range  $[d_l, d_u]$ and proportionally to the expansion of the variance in the interval $[-\sigma_\mathcal{E}, \sigma_\mathcal{E}]$:

\begin{equation}\label{eqn:deltai}
    \delta_i = \overline{\mathcal{E}} + \left(i  \frac{r}{k-1} - mp_r\right) + \left( i  \frac{2\sigma_\mathcal{E}}{k- 1} - \sigma_\mathcal{E}\right)
\end{equation}

The displacements $\delta_i$, for $i = 0 \hdots k-1$, are equally distributed in the interval $[\overline{\mathcal{E}} - d_l - \sigma_\mathcal{E}\;,\; \overline{\mathcal{E}} + d_u +  \sigma_\mathcal{E}]$ and indicate the new position of fuzzy set $A_i$, given the deviations from the known UoD, whose range is $r$ and its midpoint $mp_r$, and the error signal with the mean $\overline{\mathcal{E}}$ and variance $\sigma_\mathcal{E}$. Indeed, while the term $i(r/(k-1))-mp_r$ is used to translate the midpoint of $A_i$ by a fraction of $r$ centered in $mp_r$, the term  $ i(2\sigma_\mathcal{E}/(k-1)) -\sigma_\mathcal{E}$ is used to offset the scaling of the fuzzy set bounds by a fraction of $\sigma_\mathcal{E}$ centered in $0$.

\item [Step 4] \textit{Compute the scaling factor $\rho_i$}, Eq. \eqref{eqn:rhoi}: For each fuzzy set $A_i \in \Tilde{A}$ the scale increment is empirically calculated as the distance between the displacements $\delta_i$, in order to adjust the fuzzy set lengths proportionally to their displacements, avoiding discontinuities between the sets (intervals on $U$ not covered by any fuzzy set) by setting $l_i = c_{i-1}$ and $u_i = c_{i+1}$: 
\begin{equation} \label{eqn:rhoi}
    \rho_i = |\delta_{i-1} - \delta_{i+1}|
\end{equation}

\item [Step 5] \textit{Update $\epsilon$}: Get the last forecast value $\hat{y}(t+1)$ and the last known value $y(t+1) \in Y$. Compute the error term $\epsilon(t+1) = y(t+1) - \hat{y}(t+1)$, push it on to the residuals set $\mathcal{E}$ and delete the oldest value. That is:
\begin{equation}
    \mathcal{E} = \mathcal{E} \setminus \epsilon(t-w) \cup \epsilon(t+1) \ 
\end{equation}
\end{enumerate}
\subsection{Forecasting Procedure}
\label{sec:forecasting_procedure}

Given an input value $y(t)$, the linguistic variable $\Tilde{A}$, the inferred rule set and the perturbation parameters $\delta_i$ and $\rho_i$ for each fuzzy set $A_i \in \Tilde{A}$:

\begin{enumerate}
\item [Step 1] \textit{Fuzzification}: Compute the membership grade $\mu_{A_i}$ for each non-stationary fuzzy set $A_i$, such that $\mu_{A_i} = \mu_{A_i}(y(t), \pi(l_i, c_i, u_i, \delta_i, \rho_i))$;

\item [Step 2] \textit{Rule matching}: Build a set, $\mathcal{S}$ with all the rules $A_j \rightarrow RHS_j$ where $\mu_{A_j}(y(t)) > 0$, that is:

\begin{equation}
    \mathcal{S} = \{ A_j \rightarrow RHS_j | \mu_{A_j}(y(t)) > 0 \}
\end{equation}



\item [Step 3] \textit{Defuzzification}: Compute the forecast $\hat{y}(t+1)$ according to Equation \eqref{eqn:yhat} as the weighted sum of the rule mid-points, $mp$, by their membership grades $\mu_j$ for each selected rule $j$:

\begin{equation} \label{eqn:yhat}
\hat{y}(t+1) = \sum_{A_j \rightarrow RHS_j \in \mathcal{S}} \mu_{A_j}(y(t)) \cdot mp(RHS_j)
\end{equation}

where

\begin{equation}
   mp(RHS) =  \dfrac{\sum_{A_i \in RHS} c_{A_i}}{|RHS|}
\end{equation}

\end{enumerate}


\subsection{Method Discussion}

The NSFS has the ability to dynamically adapt  membership functions as the statistical properties of the time series vary. The NSFTS method tries to detect these changes by using the displacements of the input data $y(t)$ in relation to the known $U$ and the residuals statistics to identify bias and variance shifts. 

The major assumption of the method is that adjusting the fuzzy sets, without modifying the model structure represented by the learned rules, is enough to adapt the model to the new behavior of the time series. Such procedures, described in subsection \ref{sec:parameter_adaption_procedure}, given an input of size $T$, memory window length $W$, refreshing interval $R$ and $k$ fuzzy sets, have a time complexity of $O(T/R \cdot W \cdot \log k)$, while a retraining process would take $O(T/R \cdot W \cdot (\log k)^\Omega)$, as discussed in Section \ref{sec:timevariant}. Hence, this approach becomes much cheaper computationally, when compared to retraining a new model from scratch.

The $\pi$ function entails an additive approach for the translation and scaling of the parameters based on grid partitioning of the UoD. The procedure calculates the translation increment proportionally to the displacement of the input value in relation to the known $U$ and the variance of the residuals. The scaling increment works as a heuristic to keep the grid aspect of the fuzzy sets after the translation, connecting the lower and upper bounds with the centers of adjacent fuzzy sets.

\begin{figure*}[!htb]
    \centering
    \begin{subfloat}[UoD initial partitioning]{
        \includegraphics[width=\textwidth,height=1.8in,keepaspectratio]{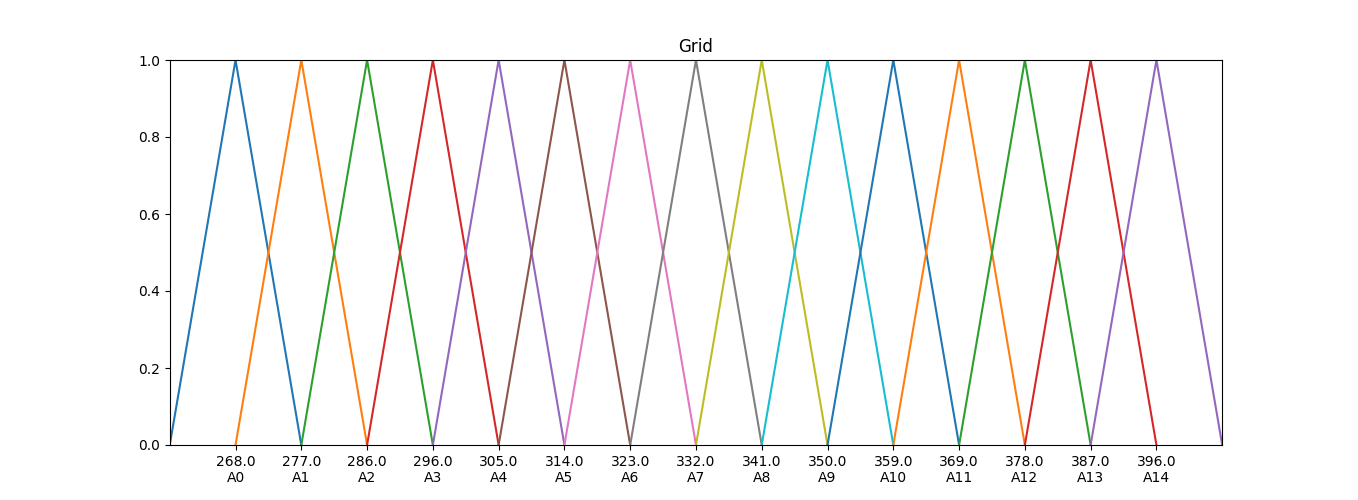}
        \label{fig:partitioning}}
    \end{subfloat}
    ~
    \begin{subfloat}[NSFS model forecasts]{
        \includegraphics[width=\textwidth,height=1.8in,keepaspectratio]{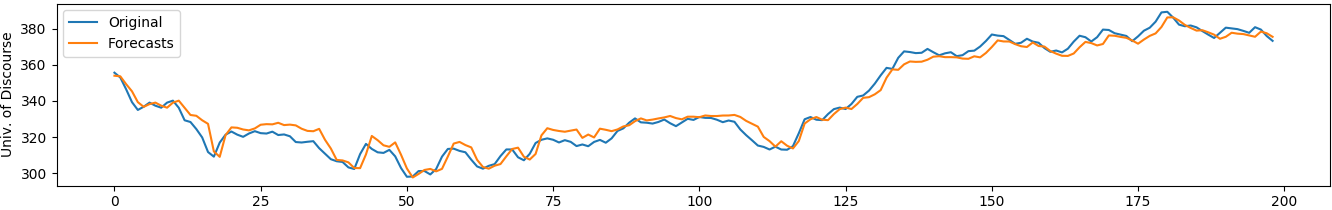}
        \label{fig:forecasts}}
    \end{subfloat}
    ~ 
    \begin{subfloat}[Residuals]{
        \includegraphics[width=\textwidth,height=1.8in,keepaspectratio]{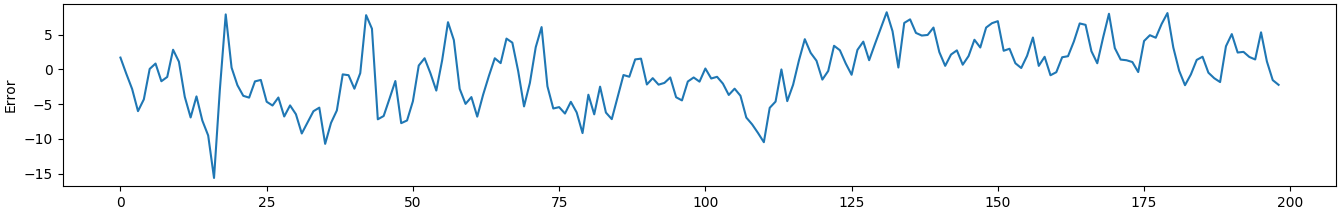}
        \label{fig:residuals}}
    \end{subfloat}
    ~ 
    \begin{subfloat}[Perturbations on the NSFS]{
        \includegraphics[width=\textwidth,height=1.8in,keepaspectratio]{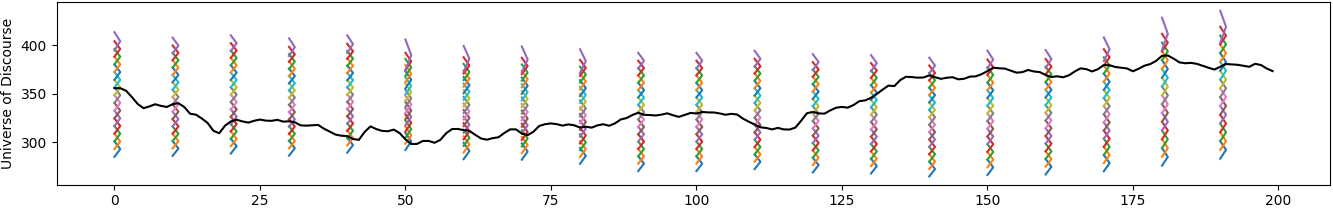}
        \label{fig:perturbations}}
    \end{subfloat}

    \caption{Out of sample example of NSFTS forecasting}
    \label{fig:nsfts}
\end{figure*}

Figure \ref{fig:nsfts} depicts the forecasting method applied to a test sample with significant drift from the training data extracted from SP500 time series, later explained in Section \ref{sec:results}. 

In Figure \ref{fig:partitioning} the partitioning of the UoD is shown with 15 partitions. The generated forecasts are presented in Figure \ref{fig:forecasts} and their residuals in Figure \ref{fig:residuals}. The perturbations on the 
NSFS, in response to the previous residuals, are represented in Figure \ref{fig:perturbations}, where the same fuzzy sets represented in Figure \ref{fig:partitioning} are now colored  over the vertical axis. It is possible to see that these fuzzy sets move, sometimes expanding sometimes contracting, in order to fit to the data. As the drift increases, the displacement and scaling of the fuzzy sets also grow.

The proposed model is a case of active learning, in which the learning algorithm is able to interactively obtain, from different sources of information, the necessary inputs for the generation or improvement of its learning. Basically, it uses residual error values to fit a predefined model. The version here presented focuses on highlighting its adaptability to changes observed over time. It uses first-order fuzzy rules, that is, considering only a previous observation, and is applied to forecasting problems in univariate time series. However, its time variant model characteristics can be expanded or adapted to other models for better prediction values. For instance, the model can be expanded to comprehend a high-order fuzzy rule generation. It also can be adapted to fit parameters of other FTS models, such as PWFTS, used as benchmark in this work.

\section{Computational Experiments}
\renewcommand{\tabcolsep}{3pt}
\label{sec:results}



\subsection{Experimental Design} 
\label{sec:experiments_design}

Different datasets were chosen for model validation. The datasets consisted of four stock market indices (Dow Jones, NASDAQ, SP500 and TAIEX), three FOREX pairs (EUR-USD, EUR-GBP, GBP-USD), two cryptocoins exchange rates (Bitcoin-USD and Ethereum-USD) illustrated in \ref{sec:appendix4} (Fig. \ref{fig:bases_stock}) and eight synthetic time series with concept drifts, see 
\ref{sec:appendix3} (Fig. \ref{fig:bases_sinteticas}).


The market indexes data sets contain the daily averaged index by business day, such that the Dow Jones Industrial Average (Dow Jones)\footnote{\url{https://finance.yahoo.com/quote/\%5EGSPC/history?p=\%5EGSPC}. Accessed in 11/11/2019} is sampled from 1985 to 2017 time window, the Taiwan Stock Exchange Capitalization Weighted Stock Index (TAIEX)\footnote{\url{https://www.twse.com.tw/en/page/products/indices/series.html}. Accessed in 11/11/2019}  is sampled from 1995 to 2014, the National Association of Securities Dealers Automated Quotations - Composite Index (NASDAQ \^IXIC)\footnote{\url{https://www.nasdaq.com/market-activity/index/ixic}. Accessed in 12/11/2019} is sampled from 2000 to 2016 and the SP500 - Standard \& Poor's 500\footnote{\url{https://finance.yahoo.com/quote/\%5EGSPC/history?p=\%5EGSPC}. Accessed in 12/11/2019} is sampled from 1950 to 2017. The FOREX data sets contain the daily averaged quotations, by business day, from 2016 to 2018, which pairs are the US Dollar to Euro (USD-EUR), Euro to Great British Pound (EUR-GBP) and Great British Pound to US Dollar (GBP-USD). The cryptocoin datasets contain the daily quotations, in US Dollar, of the Bitcoin (BTC-USD) and Ethereum (ETC-USD). The synthetic data aims to represent different types of concept drifts.

The Augmented Dickey-Fuller (ADF) test was used to check which time series are non-stationary. The  Levene’s test was used to test whether the series are heteroskedastic or not. The detailed results of these tests are shown in \ref{sec:appendix} and \ref{sec:appendix1}, respectively. In summary, except for the data sets, ``Stationary Signal'', ``Stationary Signal with blip'' and ``Sudden variance'' (see \ref{sec:appendix3} (Fig. \ref{fig:bases_sinteticas})) all the benchmarks were shown to be non-stationary and only the ``Stationary Signal with blip'' homogeneity of variances with a confidence level of 95\%.



The standard accuracy metrics used to evaluate point forecasting methods are the Root Mean Squared Error (RMSE) \eqref{eqn:rmse}, Mean Absolute Percentage Error (MAPE) \eqref{eqn:mape} and Theil's U statistic (U) \eqref{eqn:ustatistic} where $y$ are the target values, $\hat{y}$ are the forecast values and $n$ the sample size. These metrics are commonly used in evaluating forecasting models \cite{bose2019designing}. 

\begin{equation}
RMSE = \sqrt{ \frac{1}{n} \sum_{i=1}^n (y_i - \hat{y}_i)^2 }
\label{eqn:rmse}
\end{equation}

\begin{equation}
MAPE = \frac{1}{n} \sum_{i=1}^n \Big|\frac{y_i - \hat{y}}{y_i}\Big|
\label{eqn:mape}
\end{equation}

\begin{equation}
U = \frac{\sqrt{\sum_{i=1}^n (y_i - \hat{y})^2}}{\sqrt{\sum_{i=1}^n y_i^2} + \sqrt{\sum_{i=1}^n \hat{y}_i^2}}
\label{eqn:ustatistic}
\end{equation}

The proposed NSFTS was tested against the 
Time Variant \citep{song1994forecasting} and the Incremental Ensemble approaches, both using the PWFTS method \cite{Silva2019b} available in \cite{pyFTS} as internal method. As can be seen in \cite{Silva2019a}, the PWFTS outperformed a wide number of forecasting methods ranging from classic statistical ones such as ARIMA \cite{asteriou2011arima} to more recent ones such as the k-Nearest Neighbors with Kernel Density Estimation \cite{zhang2016k}. The parameters of all methods are presented in Table \ref{tab:parameters}, and they were defined through grid search.

\begin{table}[!htb]
    \centering
    \begin{tabular}{|c|c|c|}
\hline
\textbf{Method}                                                           & \textbf{Parameter} & \textbf{Value} \\ \hline
\multirow{2}{*}{All}                                                                                & k                                            & 35                                       \\ \cline{2-3} 
                                                                                                    & $\Omega$                        & 1                                        \\ \hline
\multirow{2}{*}{\begin{tabular}[c]{@{}c@{}}Song and Chissom \& \\Incremental Ensemble\end{tabular}} & W                                            & 100                                      \\ \cline{2-3} 
                                                                                                    & R                                            & 10                                       \\ \hline
Incremental Ensemble                                                                                & M                                            & 2                                        \\ \hline
\end{tabular}
    \caption{Benchmarking parameters}
    \label{tab:parameters}
\end{table}

The grid partitioning scheme was used for the initial generation of fuzzy sets, where the best number of partitions in the range $[5,100]$ was selected for each data set. The high order methods were tested with orders 2 and 3.

\subsection{Results}
\label{sec:experiments_results}

The average RMSE, MAPE and U statistic results by method and data set are presented in Table \ref{tab:experiments}. It can be seen that the NSFTS is superior or, at least, not worse than the other methods in all the selected real world benchmarks.

From the results on the artificial data sets, it can be seen that the NSFTS has more difficulties than the other methods in handling sudden changes in the series distribution mean. This behavior is expected from the NSFTS once it is an adaptive method and therefore presents a delay to respond to changes. On the other hand, it handles with relative ease incremental changes in the mean. Observing the Theil’s U statistic which is insensitive to the range of values of the series, one can see that changes in the variance alone do not seem to affect the NSFTS performance by much. 

These observations explain the good performance of the NSFTS on the economic series. As one can see in  \ref{sec:appendix4}, these series present a gradual, though significant, variation of the mean along with different types of changes in variance. Apparently, the superior performance of the NSFTS when the mean varies gradually gave it the edge over the other methods.

\begin{landscape}
\begin{table}[!htb]
\centering
\caption{Results of the metrics RMSE, MAPE and U by approach and dataset}
\label{tab:experiments}
\begin{tabular}{|l|r|r|r|r|r|r|r|r|r|}
\hline
\multicolumn{1}{|c|}{\multirow{2}{*}{Dataset}} & \multicolumn{3}{c|}{RMSE}                                                                   & \multicolumn{3}{c|}{MAPE}                                                                   & \multicolumn{3}{c|}{U}                                                                      \\ \cline{2-10} 
\multicolumn{1}{|c|}{}                         & \multicolumn{1}{c|}{T. Variant} & \multicolumn{1}{c|}{I. Ensemble} & \multicolumn{1}{c|}{NSFTS} & \multicolumn{1}{c|}{T. Variant} & \multicolumn{1}{c|}{I. Ensemble} & \multicolumn{1}{c|}{NSFTS} & \multicolumn{1}{c|}{T. Variant} & \multicolumn{1}{c|}{I. Ensemble} & \multicolumn{1}{c|}{NSFTS} \\ \hline
Stationary Signal                              & 0.07                         & 0.07                            & 0.07                       & 1.13                         & 1.16                            & 1.13                       & 0.99                         & 1.01                            & 1.01                       \\ \hline
Stationary signal with blip                    & 0.16                         & 0.17                            & 0.14                       & 1.29                         & 1.27                            & 1.22                       & 1.14                         & 1.24                            & 1.00                       \\ \hline
Sudden variance                                & 0.12                         & 0.12                            & 0.12                       & 1.80                         & 1.81                            & 1.73                       & 0.98                         & 1.00                            & 0.98                       \\ \hline
Sudden mean                                    & 0.25                         & 0.55                            & 0.79                       & 2.20                         & 6.00                            & 19.89                      & 1.72                         & 3.85                            & 5.66                       \\ \hline
Sudden mean and variance                       & 0.59                         & 0.89                            & 0.89                       & 3.82                         & 7.52                            & 19.58                      & 1.22                         & 1.90                            & 1.96                       \\ \hline
Incremental mean                               & 0.39                         & 7.81                            & 0.11                       & 1.02                         & 14.92                           & 0.69                       & 3.45                         & 71.08                           & 1.04                       \\ \hline
Incremental variance                           & 51.89                        & 50.51                           & 60.59                      & 151.43                       & 149.56                          & 241.69                     & 0.78                         & 0.78                            & 0.96                       \\ \hline
Incremental mean and variance                  & 1.29                         & 2.37                            & 1.65                       & 5.06                         & 8.48                            & 5.44                       & 1.03                         & 1.51                            & 1.09                       \\ \hline
TAIEX                                          & 145.82                       & 1130.86                         & 116.82                     & 1.27                         & 9.60                            & 1.34                       & 1.52                         & 11.90                           & 1.24                       \\ \hline
SP500                                          & 9.29                         & 61.22                           & 7.86                       & 0.64                         & 2.66                            & 0.57                       & 1.16                         & 7.73                            & 1.01                       \\ \hline
NASDAQ                                         & 35.09                        & 214.74                          & 36.75                      & 0.90                         & 4.49                            & 1.06                       & 1.25                         & 7.71                            & 1.32                       \\ \hline
DowJones                                       & 71.88                        & 519.55                          & 63.26                      & 0.64                         & 2.80                            & 0.60                       & 1.13                         & 8.25                            & 1.02                       \\ \hline
BTC-USD                                        & 465.72                       & 1775.53                         & 180.90                     & 4.97                         & 32.81                           & 3.40                       & 2.96                         & 11.50                           & 1.19                       \\ \hline
ETH-USD                                        & 44.54                        & 222.21                          & 24.16                      & 7.35                         & 41.47                           & 4.06                       & 2.06                         & 10.84                           & 1.25                       \\ \hline
EUR-USD                                        & 0.01                         & 0.04                            & 0.01                       & 0.43                         & 0.98                            & 0.41                       & 1.22                         & 6.75                            & 1.13                       \\ \hline
EUR-GBP                                        & 0.00                         & 0.01                            & 0.00                       & 0.37                         & 0.48                            & 0.32                       & 1.24                         & 3.61                            & 1.08                       \\ \hline
GBP-USD                                        & 0.01                         & 0.07                            & 0.01                       & 0.41                         & 1.18                            & 0.42                       & 1.23                         & 9.62                            & 1.26                       \\ \hline
\end{tabular}
\end{table}
\end{landscape}



To assess the overall performance of the methods with respect to the RMSE we ran a Friedman Aligned Ranks (F-test), a non-parametric test for the equality of the means was used with $\alpha = 0.05$, where $H_0$ indicates that there is no significant difference between the means and $H_1$ indicates that there is at least one significant difference between the means. The F-value result was 
$13.1994$ with a p-value of 
$0.0013$. Thus, $H_0$ was rejected at 5\% confidence level. 

A one-versus-all Finner paired \textit{post-hoc} test was employed to check the method equivalence with NSFTS as control method, where $H_0$ indicates that there is no significant difference between the means and $H_1$ indicates that there is a significant difference between the means. The results are presented in Table \ref{tab:posthoc} which shows that NSFTS is not equivalent to the Incremental Ensemble method and equivalent to the time variant FTS model, considering a 95\% confidence level. 

It is important to remind that the NSFTS is computationally cheaper than the Time Variant method since it does not require retraining. Therefore, even though they present equivalent RMSE performance, the NSFTS still has the computational edge. 

\begin{table}[!htbp]
\centering
\caption{\textit{Post hoc} multiple comparisons with the NSFTS as control method}
\label{tab:posthoc}
\begin{tabular}{|l|c|c|c|c|}
\hline
\textbf{Comparison}                                                      & \textbf{Z-value} & \textbf{P-value} & \textbf{\begin{tabular}[c]{@{}r@{}}Adjusted p-value\end{tabular}} & \textbf{Result} \\ \hline
\begin{tabular}[c]{@{}l@{}}NSFTS vs \\ Incremental Ensemble\end{tabular} & 3.680062 &  0.000233 &          0.000466 &  $H_0$ Rejected \\ \hline
\begin{tabular}[c]{@{}l@{}}NSFTS vs \\Time Variant \end{tabular}      & 0.144203 &  0.885340 &          0.885340 &  $H_0$ Accepted\\ \hline
\end{tabular}
\end{table}

\section{Conclusion}
\label{sec:conclusion}

This paper proposed a Non-Stationary Fuzzy Time Series (NSFTS) method that is able to dynamically adapt its fuzzy sets to reflect the changes in the underlying stochastic processes based on the residual errors. The proposed approach enables the model to be trained only once and remain useful long after. 

The parameter adaptation procedure developed here can be integrated into other FTS methods, extending them to deal with forecasting in non-stationary environments.

The method was tested with several non-stationary and heteroskedastic data sets consisting of four market indices, three FOREX pairs, two cryptocoin exchange rates and eight synthetic time-series that present different combinations of concept-drifts.  

The main feature delivered by the proposed method is the capability of capturing a wide spectrum of unconditional heteroskedastic effects and time series trends by the variation of several parameters of its internal model. This is done without data pre-processing, without need of retraining and without changing the symbolic structure in the learned rules. 
In order to contribute to the replication of all the results in the paper, we provide full results and all source codes for this article in Github and in the MINDS Laboratory website. The link is \url{https://github.com/PYFTS/NSFTS}.

\section*{Acknowledgements}

This work was partially financed by the Foundation for Research of the State of Minas Gerais (Fundação de Amparo à Pesquisa do Estado de Minas Gerais - FAPEMIG) and by the National Council for Scientific and Technological Development (CNPq), Brazil, Grants no. 405911/2017-3; no. 169392/2017-1; and no. 306850/2016-8.

\bibliography{mybibfile}

\appendix

\clearpage
\section{Appendix}
\label{sec:appendix4}

\ref{sec:appendix4}: Stock market indices

\begin{figure}[!htb]
    \centering
    \includegraphics[width=1\textwidth]{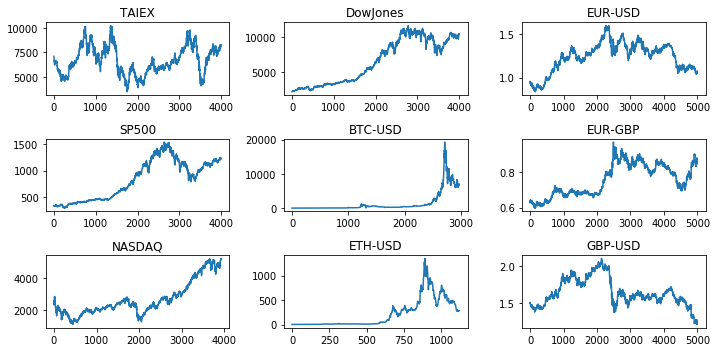}
    \caption{Stock market indices (TAIEX, Dow Jones, NASDAQ and SP500), FOREX pairs (EUR-USD, EUR-GBP, GBP-USD) and cryptocoin exchange rates (Bitcoin-USD and Ethereum-USD).}
    \label{fig:bases_stock}
\end{figure}

\newpage
\section{Appendix}
\label{sec:appendix3}

\ref{sec:appendix3}: Synthetic data sets

\begin{figure}[!htb]
    \centering
    \includegraphics[width=1\textwidth]{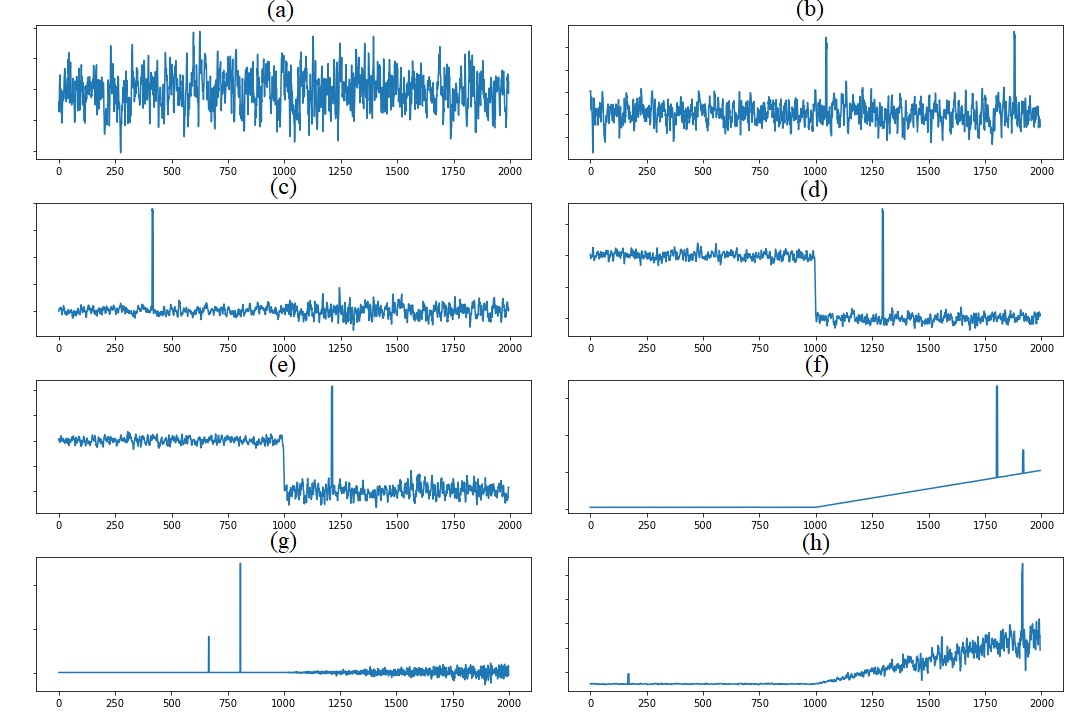}
    \caption{Synthetic time series with different combinations of concept drifts. These are: (a) stationary signal; (b) stationary signal with blip; (c) sudden variance; (d) sudden mean; (e) sudden mean and variance; (f) incremental mean; (g) incremental variance; (h) incremental mean and variance}.
    \label{fig:bases_sinteticas}
\end{figure}

\newpage

\section{Appendix}
\label{sec:appendix}

In order to evaluate the stationarity of the presented data sets, the Augmented Dickey-Fuller (ADF) test for unit root was employed considering a confidence level $\alpha = 0.05$, where $H_0$ indicates that the time series have a unit root and it is non-stationary and $H_1$ indicates that time series does not have a unit root and it is stationary. The Augmented Dickey-Fuller results for unit root are shown below.

\begin{table}[!htb]
\centering
\caption{Stationarity evalution}
\label{tab:datasets}
\begin{tabular}{lrrc}
\hline
\multicolumn{1}{c}{\multirow{2}{*}{Dataset}} & \multicolumn{3}{c}{Augmented Dickey-Fuller}                                \\ \cline{2-4} 
\multicolumn{1}{c}{}                         & \multicolumn{1}{c}{Statistic} & \multicolumn{1}{c}{p-value} & $H_0$ result \\ \hline
Stationary Signal                            & -7.427114                     & 6.504708e-11                & Rejected     \\
Stationary signal with blip                  & -7.497758                     & 4.334045e-11                & Rejected     \\
Sudden variance                              & -7.746345                     & 1.029561e-11                & Rejected     \\
Sudden mean                                  & -2.112067                     & 2.396902e-01                & Accepted     \\
Sudden mean and variance                     & -1.165176                     & 6.883938e-01                & Accepted     \\
Incremental mean                             & 3.286850                      & 1.000000e+00                & Accepted     \\
Incremental variance                         & -24.746787                    & 0.000000e+00                & Rejected     \\
Incremental mean and variance                & -2.217183                     & 2.000681e-01                & Accepted     \\
TAIEX                                        & -2.491767                     & 1.174904e-01                & Accepted     \\
SP500                                        & -0.943446                     & 7.733287e-01                & Accepted     \\
NASDAQ                                       & 0.476224                      & 9.841323e-01                & Accepted     \\
DowJones                                     & -0.800893                     & 8.188597e-01                & Accepted     \\
BTC-USD                                      & -1.206100                     & 6.709662e-01                & Accepted     \\
ETH-USD                                      & -1.852677                     & 3.546403e-01                & Accepted     \\
EUR-USD                                      & -1.845986                     & 3.578816e-01                & Accepted     \\
EUR-GBP                                      & -1.845986                     & 3.578816e-01                & Accepted     \\
GBP-USD                                      & -1.131750                     & 7.022457e-01                & Accepted     \\ \hline
\end{tabular}
\end{table}

\newpage

\section{Appendix}
\label{sec:appendix1}

To check the homoskedasticity the Levene’s test was employed, which checks for homogeneity of variances, with confidence level $\alpha = 0.05$, where $H_0$ indicates that the sub-samples variances of the time series are all equal and $H_1$ indicates that at least one variance of the time series sub-samples is different. the Levene’s results for homogeneity of variances are shown below.

\begin{table}[!htb]
\centering
\caption{Homogeneity of variances evaluation}
\label{tab:datasets}
\begin{tabular}{lrrc}
\hline
\multicolumn{1}{c}{\multirow{2}{*}{Dataset}} & \multicolumn{3}{c}{Levene’s test}                                          \\ \cline{2-4} 
\multicolumn{1}{c}{}                         & \multicolumn{1}{c}{Statistic} & \multicolumn{1}{c}{p-value} & $H_0$ result \\ \hline
Stationary Signal                            & 10.148665                     & 1.466392e-03                & Rejected     \\
Stationary signal with blip                  & 1.935753                      & 1.642857e-01                & Accepted     \\
Sudden variance                              & 25.163718                     & 5.733279e-07                & Rejected     \\
Sudden mean                                  & 6.989502                      & 8.263198e-03                & Rejected     \\
Sudden mean and variance                     & 173.802832                    & 4.097776e-38                & Rejected     \\
Incremental mean                             & 2954.661000                   & 0.000000e+00                & Rejected     \\
Incremental variance                         & 520.174358                    & 1.597361e-102               & Rejected     \\
Incremental mean and variance                & 521.736508                    & 9.142922e-103               & Rejected     \\
TAIEX                                        & 62.530885                     & 3.366934e-15                & Rejected     \\
SP500                                        & 64.954050                     & 1.003282e-15                & Rejected     \\
NASDAQ                                       & 1851.123985                   & 0.000000e+00                & Rejected     \\
DowJones                                     & 163.413938                    & 1.053157e-36                & Rejected     \\
BTC-USD                                      & 524.853179                    & 4.352254e-107               & Rejected     \\
ETH-USD                                      & 666.975156                    & 9.700699e-116               & Rejected     \\
EUR-USD                                      & 401.104689                    & 6.851098e-86                & Rejected     \\
EUR-GBP                                      & 401.104689                    & 6.851098e-86                & Rejected     \\
GBP-USD                                      & 1340.896808                   & 2.812862e-260               & Rejected     \\ \hline
\end{tabular}
\end{table}

\end{document}